\documentclass{article} 
\usepackage[preprint]{colm2025_conference}

\usepackage{microtype}
\usepackage{hyperref}
\usepackage{url}
\usepackage{booktabs}

\usepackage{lineno}
\usepackage{enumitem}
\usepackage{wrapfig}
\usepackage{graphicx,subcaption}
\usepackage{caption}
\let\cite\citep

\usepackage{color, soul}
\usepackage{svg}
\usepackage{graphicx}
\usepackage{amsmath}
\usepackage{algorithm}
\usepackage{algpseudocode}
\usepackage{listings}
\usepackage{amssymb}
\usepackage{multirow}
\newcommand{\ada}{text-ada-001}
\newcommand{\babbage}{text-babbage-001}
\newcommand{\curie}{text-curie-001}
\newcommand{\davinci}{text-davinci-002}
\newcommand{\llama}{Llama}
\newcommand{\gemma}{Gemma}
\newcommand{\mistral}{Mistral}
\newcommand{\qwen}{Qwen}
\newcommand{\ours}{MetaLLM}

\definecolor{darkblue}{rgb}{0, 0, 0.5}
\definecolor{DarkGreen}{rgb}{0.0, 0.5, 0.0}
\hypersetup{colorlinks=true, citecolor=darkblue, linkcolor=darkblue, urlcolor=darkblue}

\title{MetaLLM: A High-performant and Cost-efficient Dynamic Framework for Wrapping LLMs}

\author{%
  \parbox{.85\linewidth}{\centering Quang H. Nguyen$^1$, Thinh Dao$^1$, Duy C. Hoang$^1$, Juliette Decugis$^1$, \\Saurav Manchanda$^{2}$, Nitesh V. Chawla$^{3}$, Khoa D. Doan$^1$} \\
$^1$College of Engineering and Computer Science, VinUniversity, Vietnam \\
$^2$Amazon, USA\thanks{The work does not relate to the author's position at Amazon.}\\
$^3$University of Notre Dame\\
\texttt{\{quang.nh, 21thinh.dd, duy.hc, juliette.d\}@vinuni.edu.vn, }\\ \texttt{sauravm.kgp@gmail.com, nchawla@nd.edu, khoa.dd@vinuni.edu.vn}
}
\begin{document}

\ifcolmsubmission
\linenumbers
\fi

\maketitle

\begin{abstract}
The rapid progress in machine learning (ML) has brought forth many large language models (LLMs) that excel in various tasks and areas. These LLMs come with different abilities and costs in terms of computation or pricing. Since the demand for each query can vary, e.g., because of the queried domain or its complexity, defaulting to one LLM in an application is not usually the best choice, whether it is the biggest, priciest, or even the one with the best average test performance. Consequently, picking the right LLM that is both accurate and cost-effective for an application is necessary yet remains a challenge. In this paper, we introduce MetaLLM, a framework that dynamically and intelligently routes each query to the optimal LLM (among several available LLMs) for classification and multi-choice question-answering tasks, achieving significantly improved accuracy and cost-effectiveness. By framing the selection problem as a multi-armed bandit, MetaLLM balances prediction accuracy and cost efficiency under uncertainty. Our experiments, conducted on popular LLM platforms such as OpenAI and Together AI, as well as open-source LLM, showcase MetaLLM's efficacy in real-world scenarios, laying the groundwork for future extensions\footnote{Our code is available \href{https://github.com/mail-research/MetaLLM-wrapper/}{here}.}.
\end{abstract}

\section{Introduction}
Large language models have shown extraordinary zero-shot capabilities in various tasks and domains, such as text classification, summarization, question answering, and chatbot~\cite{radford2019language, brown2020language, schick2021exploiting, wei2022chain, ouyang2022training, trivedi2023interleaving}. 
Recent works~\cite{kaplan2020scaling, chowdhery2023palm, hoffmann2022training} suggest exhaustively scaling the model size and training data size to improve the performance of language models and provoke their emergent abilities; for example, GPT-4, with over $1.74$ trillion parameters, achieves superior performance in several tasks but also incurs high economic costs. While this trend of scaling language models will continue in the future, there is also a growing diversification in recent models in terms of task or (sub-)domain specialization and computational costs. 
As a consequence, no single model --even the largest and most expensive -- can consistently yield the best performance across all tasks.
This means that, for model users, identifying which LLM is best suited for their applications will become crucial. However, this is a challenging task, especially when we factor in cost constraints, either in terms of computational resources or API service pricing. 


We imagine a world with several LLM providers, such as OpenAI\footnote{\url{https://platform.openai.com/docs/models}} or Together AI\footnote{\url{https://www.together.ai/}}; each provides service access to a diverse suite of LLMs with heterogeneous capabilities and cost structures. 
Here, an LLM user asks this important question: \textit{How do I select an LLM $i$ (out of $k$ LLMs) for optimal performance and usage cost in my application?} 
One option is \textit{combining multiple LLMs} as seen in existing ensemble methods~\cite{jiang2023llm, wang2023fusing, ong2024routellm}, but this approach will yield significantly higher service costs; another option is cascading over a set of LLMs~\cite{chen2023frugalgpt}, but such an approach still requires querying the LLMs until we could find the best one. On the other hand, defaulting to a single LLM to avoid the extra cost of querying multiple models -- by predicting the performance of LLMs~\cite{shnitzer2023large, lu2023routing} -- may not also be optimal; for some queries, a less expensive model may also provide correct answers.  
Furthermore, as different LLMs exhibit very distinctive abilities on different tasks and data distribution~\cite{jiang2023llm, wang2023fusing}, for a query, it is possible for LLM $i$ to perform better than LLM $j$ even though the cost of LLM $i$ is noticeably lower than the cost of $j$. 

In this work, we introduce a framework that wraps around a set of LLMs with diverse capabilities and brings the best performance at a more affordable cost to a user's application. The underlying component of this framework is a multi-arm bandit algorithm that routes each query to the least expensive LLM with the correct answer for the zero-shot classification and question answering task. Essentially, we formulate the problem of selecting the ``best'' LLM for a query as a decision-making problem under uncertainty: when a new query arrives, the proposed framework picks an LLM to answer and observe whether the LLM provides a correct answer; the goal is to maximize the total reward -- a specific trade-off between the overall performance and usage cost. Intuitively, our method prefers the LLM that has a high probability of answering a query at a low cost.
Our contributions can be summarized as follows:

\begin{itemize}[leftmargin=*]
    \item We propose \ours, a versatile wrapper around a suite of any off-the-shelf LLMs, for zero-shot text classification tasks. \ours{} can intelligently choose the target LLM for each query to achieve optimal performance and cost. 
    \item We propose an algorithm based on multi-armed bandit to tackle the routing problem in \ours. This algorithm is efficient since it makes the routing decision without needing to query any LLMs. 
    \item Our experimental results on benchmark datasets and popular API services, including OpenAI and Together AI, and both closed-source and open-source LLMs,  demonstrate the ability of \ours~in identifying the optimal LLM in terms of cost and performance. Specifically, \ours~improves the accuracy of the best model by around $1\%$ while saving up to $60\%$ and $10\%$ of the total price on OpenAI and Together AI APIs, respectively.
\end{itemize}


Our work focuses on zero-shot classification and multiple-choice question-answering problems, which are common evaluation choices in the related work~\cite{ong2024routellm}. Nevertheless, the MetaLLM framework can be extended to arbitrary language tasks by modifying the reward function to incorporate suitable metrics assessing the quality of the responses.
We leave it for future work. 

 

\section{Related Works}
\noindent\textbf{Large Language Models.} 
The emergence of large language models (LLMs) has fundamentally transformed several domains, including natural language processing, computer vision, and e-commerce, and diverse tasks such as (zero-shot) classification, question-answering, and recommendation~\cite{menghani2023efficient,liu2023apre,liu2023bpre}. The impressive effectiveness and versatility of these LLMs have come at the price of a drastic increase in LLM sizes, along with significant computational costs and data required to train, and expensive computational resources to perform inference with them. Consequently, several companies or services are now offering users access to LLMs with diverse capabilities and heterogeneous cost structures. For instance, the costs for processing 10 million tokens are \$0.1, \$0.18, \$0.2, and \$0.3 for \gemma, \llama, \mistral, and \qwen, respectively, using Together AI APIs. This abundance of API choices significantly burdens their users with the decision ``\textit{which LLM should I use?}'', as different LLM APIs are known for their diverse capabilities for the prediction tasks~\cite{liang2023holistic,mckenzieinverse,mckenzie2022inverse}.




\noindent\textbf{Prompt Optimization and Mixture of Experts.} Fine-tuning is a standard option to improve the performance of LLMs for specific tasks. 
Mixture-of-Experts (MoE)~\cite{eigen2013learning,shazeer2017outrageously,du2022glam,si2023getting} trains a routing operator within the large model to enhance its performance; essentially, MoE assumes the model as a collection of ``experts'' (modules) and learn to route the input to the best expert. 
These approaches require training the LLMs, which is challenging for many users, while their single-LLM enhancements are usually model and scenario specific. Prompting reasoning like Chain-of-Thought ~\cite{wei2022chain,wang2022self,zhou2022least} or Tree of Thoughts~\cite{yao2024tree} could improve the LLM performance without additional training. Both MoE and prompt-based reasoning, however, could not benefit from the large number of available LLMs, some of which could be significantly less expensive to use.

\noindent\textbf{Model Ensemble.} LLMs have been shown to yield diverse capabilities due to their architecture and dataset~\cite{jiang2023llm, wang2023fusing}. \citet{jiang2023llm} observe that over $5000$ instructions, the optimal LLM for each query significantly varies, and there is no single optimal model regardless of their size. Therefore they recommend ensembling to combine the strength of multiple LLMs for a better performance. 
FrugalML~\cite{chen2020frugalml, chen2022efficient} cascades multiple machine learning models by querying sequentially until getting a response with a high confidence score. 
Motivated by FrugalML, \citet{chen2023frugalgpt} exploit the full potential of LLMs by applying multiple techniques, including prompt engineering, caching, and cascading, for higher-quality answers. 

\noindent\textbf{Model Selection.} Different from previous approaches that combine the strength of multiple models, there are some attempts to query a single LLM for each task. \citet{hari2023tryage}
suggest training a language model to predict the performance of LLMs and route to the model with the highest performance, which incorporates the high cost of the router. \citet{vsakota2024fly} formulate the cost-performance trade-off as an integer linear programming (ILP) problem and use existing ILP solvers to assign appropriate LLMs for each input query. \citet{lu2023routing} distill the preference of off-the-shelf reward models to select the LLM that achieves the best performance, ignoring its cost. \citet{ding2024hybrid} train a router that maps easy queries to small models and hard queries to large models; however, their framework is only applicable when we have two LLMs. Recently, \citet{hu2024routerbench} proposed a benchmark to evaluate routing models by their cost and performance on downstream tasks.

\noindent\textbf{Commercialized Foundation Model Services.} Recently, many companies have commercialized their LLM models so that customers can apply or even fine-tune LLMs for their own use cases without heavy technical knowledge. In 2023, OpenAI released many language models, such as \ada, \babbage, \curie, and \davinci\footnote{These models have been deprecated since 2024-01-04.}, with various prices and capabilities. Their users can use these models as a chatbot, extract text embedding, or even ask them to write code. 
Recently, Together AI also brought forth their APIs that allow access to several foundation generative models, including text generation, image generation, and multimodal models, such as \llama, \gemma, \mistral, \qwen, Flux, etc. 

\section{MetaLLM Framework}
\begin{figure}
    \centering
    \includegraphics[width=1.0\textwidth]{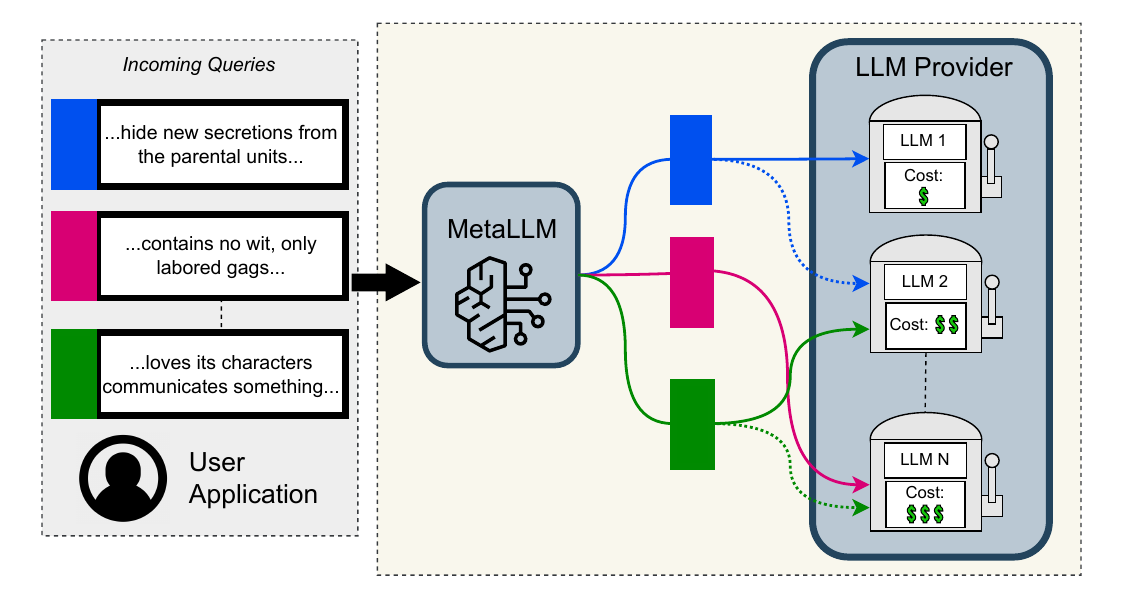}
    \caption{\textbf{The general process of serving queries in  \ours}. \ours~ wraps around an existing LLM Provider, inspects each query, and then routes it to the least expensive LLM that can provide an accurate response. As an example, the \textcolor{blue}{\textbf{blue query}} can be answered accurately by LLM 1 and LLM 2, but \ours~will route it to LLM 1 since it is less expensive; similarly in another example, the \textcolor{DarkGreen}{\textbf{green query}} is routed to the least expensive LLM 2 even though LLMs 2 and 3 both can accurately answer it. The entire process is lightweight and can be performed without accessing in LLMs.}
    \label{fig:framework}
\end{figure}
\subsection{Preliminaries}
\noindent\textbf{Zero-shot Classification.} In this paper, we employ LLMs for text classifications without additional training. Given an input sentence $x\in \mathcal{X}$, we create a prompt to ask a language model $M$ for a label.
The model gives a correct prediction if the answer $M(x)$ matches the corresponding ground-truth label $y$.

\noindent\textbf{LLM APIs.} We consider the case where the user has access to a set $K$ of $k$ different LLM APIs; each LLM $M_i$ has a different capability, determined by how well $M_i$ can answer a query $x$, and a cost $c_i$. 
For zero-shot classification, we represent the capability $a_i(x)\in \{0, 1\}$ of an LLM $M_i$ on a sample $x$ by comparing the answer to its corresponding ground-truth label: $a_i(x)=1$ if $M_i(x) = y$. 
Usually, the model with a higher cost has better capability. If the user has a limited budget, they can choose less expensive models for their application. In contrast, if they require better performance, more expensive models will be more desirable. However, in general, the more expensive model is \textit{not always} the best-performing choice for all queries~\cite{jiang2023llm, wang2023fusing}, making the LLM selection problem significantly challenging.

\noindent\textbf{Problem Formulation.} The goal of \ours~ is to learn a routing function $f:\mathcal{X}\to K$ that dispatches a query $x$ to an appropriate LLM to achieve a good response with a lower cost. For example, given a subset of LLMs, $K' \subseteq K$, that can give good responses for $x$, \ours's objective is to return $\arg\min_{l \in K'} c_{l}$.
In practice, the user wants the ability to balance the performance and the usage cost, depending on the needs of their application. 
More specifically, given a budget $b$, the user wants to maximize the performance on $N$ queries while spending less than $b$ to query the APIs. 
\begin{align}\label{formula:main_obj}
    \arg\max_f \sum_{i=1}^N a_{f(x_i)}(x_i)  \quad \text{s.t.}\quad\sum_{i=1}^N c_{f(x_i)} \leq b 
\end{align}

\subsection{The optimal solution of routing objective}
Let $S\in\{0, 1\}^{N\times k}$, $S_{i,j}=\mathbf{1}[f(x_i)=j]$, represent the choice of the routing function $f$ on $N$ queries, and
$A\in\mathbb{R}^{N\times k}$, $A_{i,j}=a_j(x_i)$, represent the performance of the $j$-th API on each sample $x_i$. Eqn.~\eqref{formula:main_obj} aims to choose the highest performance LLM within a cost constraint $b$ and can be reformulated as follows:
\begin{align}\label{formula:linear_form}
    \arg\max_S \sum_{i=1}^N\sum_{j=1}^k A_{i,j}S_{i,j}   \quad \text{s.t.}\quad\sum_{i=1}^N c_{f(x_i)} \leq b. 
\end{align}
Instead of solving a discrete optimization problem, we relax $S\in\{0, 1\}^{N\times k}$ to $S\in\mathbb{R}^{N\times k}$, $\sum_{j=1}^k S_{i,j}=1$, and has the following dual problem~\cite{Boyd_Vandenberghe_2004}:
\begin{align}\label{formula:dual_form}
    \arg\min_{p\in\mathbb{R}, q\in\mathbb{R}^N} p + \sum_i q_i  \quad \text{s.t.}\quad pc_j + q_i \geq a_j(x_i) + 1. 
\end{align}
\citet{chen2022efficient} study a similar formulation for model cascading and suggest solving the dual form as in Eqn.~\eqref{formula:dual_form}. In the case where we have an exact accuracy matrix $A$, we can optimize Eqn.~\eqref{formula:linear_form} to find the optimal $S$  and choose the suitable API $j$ such that $S_{i,j}=1$.

\subsection{The Proposed \ours}

\begin{wrapfigure}{r}{0.5\textwidth}
\vspace{-20pt}

\begin{minipage}{0.5\textwidth}
\begin{algorithm}[H]
\caption{\ours~framework}
\label{algo:main}
\begin{algorithmic}[]

\item{\textbf{input:}} $k$ LLMs, each with cost $c_j$ and accuracy $a_j(\cdot)$; cost scaling $p$, parameters $\theta$, training sample $\{x_1, \dots, x_n\}$, test query $x$.

\item{\textbf{output:}} The optimal LLM for the query $x$.
\\
\Statex \textit{\#\#\# Initialize the policy}

\For {j in 1..k}
    \State{$A_j\gets(X^{\intercal}X + \lambda I)$, $b_j\gets X^{\intercal}y_j$}
    \State{$\theta^0_j \gets A_j ^{-1} b_j$}
\EndFor
\\
\Statex\textit{\#\#\# Inference}
\State{$j\gets \arg\max_{j'} x^{\intercal}\theta_{j'} + \alpha \sqrt{x^{\intercal}A_{j'}^{-1}x}$}
\State{$A_j\gets A_j+xx^{\intercal}$, $b_j\gets b_j + r(x,j)x$}
\State{$\theta_j\gets A_j^{-1}b_j$}

{\parindent0pt\Return j}
\end{algorithmic}

\end{algorithm}
\end{minipage}
\end{wrapfigure}

When deploying an application, we cannot know the exact accuracy of an LLM's API on a test sample before sending the test sample through this LLM. 
Previous works~\cite{chen2022efficient, chen2020frugalml} learn a model to predict the performance of each API. These methods cascade multiple machine learning models and query them iteratively until the response has high confidence; consequently, they are highly expensive, especially when there are a substantial number of queries.

In this work, we approach this problem from a different perspective. Instead of training an accuracy predictor, we reformulate this problem as a multi-armed bandit. Specifically, for each input query, we define an LLM as an ``arm'', and obtain a reward expressing the performance of the LLM and the cost for that query. The benefit of this formulation is twofold: first, it allows the modeler to focus on designing the reward function to capture their application needs, making the framework more versatile; second, we can take advantage of the extensive and well-developed research on multi-arm bandits, including their rigorous theoretical foundations and practical solutions. The \ours{} framework is depicted in Figure~\ref{fig:framework}.


\noindent\textbf{Reward function.} The remaining question will discuss our design of the reward function that takes the input query and returns the reward of choosing the $i$-LLM. \citet{chen2022efficient} prove that if a cost scaling $p\in\mathbb{R}$ is the solution of Eqn.~\eqref{formula:dual_form}, the routing function $f(x)=\arg\max_i a_i(x)-pc_i$ will be the optimal solution of Eqn.~\eqref{formula:main_obj}. 
Intuitively, this strategy prefers the LLM with high performance and low-cost value.
Motivated by that theoretical result, we propose the following reward function for training the multi-armed bandit:
\begin{equation}\label{formula:reward}
    r(x, i) = a_i(x) - pc_i.
\end{equation}

\noindent\textbf{Multi-arm bandit.} We assume that the user has access to a training dataset of $n$ queries and can obtain the training performance of each LLM on this dataset.
We initialize the reward model $Q_{j}(x;\theta)$ of each arm $j$ by minimizing the objective
$\theta_j = \arg\min_{\theta}\|Q_j(x_i,\theta) - (a_j(x_i) - p c_j)\|_2^2.$

Similar to~\cite{li2010contextual}, we employ a linear reward model $Q_{j}(x;\theta_j)=x^{\intercal}\theta_j$ and obtain the closed-form solution of ridge regression $\theta^0_j = \big({A^0_j}\big)^{-1}b^0_j, A^0_j=(X^{\intercal}X + \lambda I), b^0_j= X^{\intercal}y_j$,
where $X$ is the data matrix and $y_{j,i}=a_j(x_i) - pc_j$. 

Given a new query $x_t$, we choose an arm using the UCB selection strategy:
\begin{equation}
    j=\arg\max_{j'} x_t^{\intercal}\theta^t_{j'} + \alpha \sqrt{x_t^{\intercal}{A^t_{j'}}^{-1}x_t}, \quad\alpha>0.\nonumber
\end{equation}
and update the reward function of the chosen arm, as follows:
\begin{equation}    \theta^{t+1}_j=\big({A^{t+1}_{j}}\big)^{-1}b^{t+1}_j, A^{t+1}_j = A^{t}_j + x_tx_t^{\intercal}, b^{t+1}_j=b^t_j+r(x_t,j)x_t. \nonumber
\end{equation}







\section{Experiments} \label{sec:experiments}
In this section, we provide empirical results of \ours~on popular APIs and benchmark datasets.
\subsection{Experimental Setup}

\paragraph{LLM Services.} We conduct our experiments with LLMs provided by popular API services, including OpenAI and Together AI. We chose four models from OpenAI: \ada, \babbage, \curie, and \davinci. These models have different costs and different capabilities, giving the users diverse options. The costs of these models are shown in Table~\ref{tab:openai_cost}.

    
Given a sample \verb|SENT|, we query OpenAI models with the following prompt: 
\begin{lstlisting}
For the sentence: SENT, is the sentiment in this sentence positive or negative?
\end{lstlisting}

For Together AI APIs, we evaluate \ours~with four different LLMs: \gemma, \llama, \mistral, and \qwen. Since these models come from different families, this evaluation setting exhibits more heterogeneity than the setting with the OpenAI models. We summarize the cost of each APIs in Table~\ref{tab:together_cost}.
    

\begin{table}[ht!]
    \centering
    \begin{minipage}{0.45\textwidth}
        \centering
    \begin{tabular}{lc}
    \toprule
    Model & Cost \\
    \midrule
       \ada  & \$0.40 / 1M tokens \\
        \babbage & \$0.50 / 1M tokens \\
        \curie & \$2.00 / 1M tokens \\
        \davinci & \$20.00 / 1M tokens \\
        \bottomrule
    \end{tabular}
    \caption{Price of OpenAI APIs.}    
    \label{tab:openai_cost}    
    \end{minipage}
    \hfill
    \begin{minipage}{0.45\textwidth}
        \centering
    \begin{tabular}{lc}
    \toprule
    Model & Cost \\
    \midrule
        \gemma & \$0.10 / 1M tokens \\
       \llama  & \$0.18 / 1M tokens \\
        \mistral & \$0.20 / 1M tokens \\
        \qwen & \$0.30 / 1M tokens \\
        \bottomrule
    \end{tabular}
    \caption{Price of Together AI APIs.}
    \label{tab:together_cost}    
    \end{minipage}
\end{table}

    

Given a sample \verb|SENT|, we query Together AI APIs with the following prompt: 
\begin{lstlisting}
Answer the following multiple-choice question with only the answer letter (A, B, C, or D):
Question: QUESTION
Options: 
A. CHOICE A
B. CHOICE B
C. CHOICE C
D. CHOICE D
Answer:
\end{lstlisting}

\paragraph{Datasets.}
We conduct experiments on SST-2 -- a text classification dataset and MMLU -- a question answering dataset. 
SST-2 is a binary sentiment analysis dataset consisting of movie reviews; the task here is to classify whether a review is positive or negative. This dataset has $67,439$ training samples and $872$ test samples.
MMLU is a dataset multitask language understanding
dataset including 57 multi-choice question
answering subtasks. It has $14,042$ test samples and $99,842$ auxiliary training samples collected from several question answering benchmarks.



\begin{table}[ht!]
\centering

\begin{tabular}{llcc}
\toprule
Setting & Method     & Test cost & Test accuracy  \\ \midrule
&\ada          & 0.096   & 80.50              \\
&\babbage      & 0.120   & 82.80          \\

&\curie        & 0.480   & 90.60            \\
&\davinci      & 4.800   & 91.51              \\
\midrule
\multirow{3}{*}{Offline} &\ours~(\babbage~budget) & 0.120   & 84.06           \\ 
&\ours~(\curie~budget)    & 0.539     & 89.56    \\
&\ours~(\davinci~budget)   & 2.030   & 91.97              \\ 
\midrule
\multirow{3}{*}{Online w/o training data} &\ours~(\babbage~budget) & 0.303   & 84.06           \\
&\ours~(\curie~budget) & 0.818   &     87.27     \\
&\ours~(\davinci~budget)   & 1.456   & 88.30  
\\ 
\midrule
\multirow{3}{*}{Online} &\ours~(\babbage~budget) & 0.117   & 84.06           \\
&\ours~(\curie~budget) & 0.455   & 90.94         \\
&\ours~(\davinci~budget)   & 2.046   & 92.55  \\ 

\bottomrule
\end{tabular}

\caption{The cost and accuracy of each LLM on OpenAI APIs}
\label{tab:openai_results}
\end{table}

\textbf{Training \ours.} For each input query, we utilize Sentence-BERT~\cite{reimers2019sentence} model to extract an embedding vector. \ours~is a linear model that maps the embedding vector to the reward expectation, which is optimized with the true reward (Eqn.~\eqref{formula:reward}) by Algorithm~\ref{algo:main}. We normalize the cost of each LLM in the reward function such that the highest value is $1$. For a budget $b$, we train \ours~with the scaling $p$ five times, such that the cost of \ours~on the validation set is not higher than $b$, and compute the accuracy of the classification task. 
 Given that $p$, we evaluate three different scenarios: i) only optimizing the reward function with training data, ii) only performing online updates during inference, and iii) initializing the policy from training data and updating during inference.


\textbf{Evaluation.} To evaluate, we compute the cost and the accuracy of \ours~on the test set of the classification task and compare them to each LLM candidate. 
We report the average cost per $10,000$ queries in all experiments.

\subsection{Performance of \ours~on OpenAI Models}
Table~\ref{tab:openai_results} shows the accuracy and cost of each OpenAI LLM and \ours~with different $p$ on SST-2. As can be observed, the LLM with a higher cost has better accuracy. Noticeably, the differences in cost between LLMs can be very high; for example, \davinci~is ten times more expensive than \curie~but only has $0.4\%$ better performance in terms of accuracy. Consequently, querying only the most expensive LLM is not an optimal choice unless the usage budget of the application is sufficiently high.

\noindent \textbf{Performance of \ours.} For each LLM, except \ada, we compute the cost of choosing that model only,
fine-tune the scaling $p$ on the validation set such that the cost is not higher than that model's cost (the budget),
train \ours~with the found $p$ five times and report the mean accuracy of the text classification task with LLMs selected by \ours.
We can observe that \ours~ can achieve better accuracy at a lower cost. Specifically, with the same budget as \babbage, \ours's performance is $~2\%$ better than that of defaulting all queries to \babbage, while having a slightly less expensive cost. 
Similarly, \ours~can achieve higher accuracy and lower cost than \curie.
More importantly, \ours{} can reach a higher performance than that of defaulting to \davinci, the best LLM, while spending $60\%$ less. 

\noindent\textbf{Comparison of different scenarios.} Table~\ref{tab:openai_results} also demonstrates that only optimizing on training data can yield higher accuracy than \davinci, yet not as effective as online updating on test data. The initialization step also plays an important role; only online updating fails to find a good policy. 

\begin{figure}[ht!]
    \centering
    \begin{minipage}[b]{0.55\textwidth}
        \includegraphics[width=1.0\textwidth]{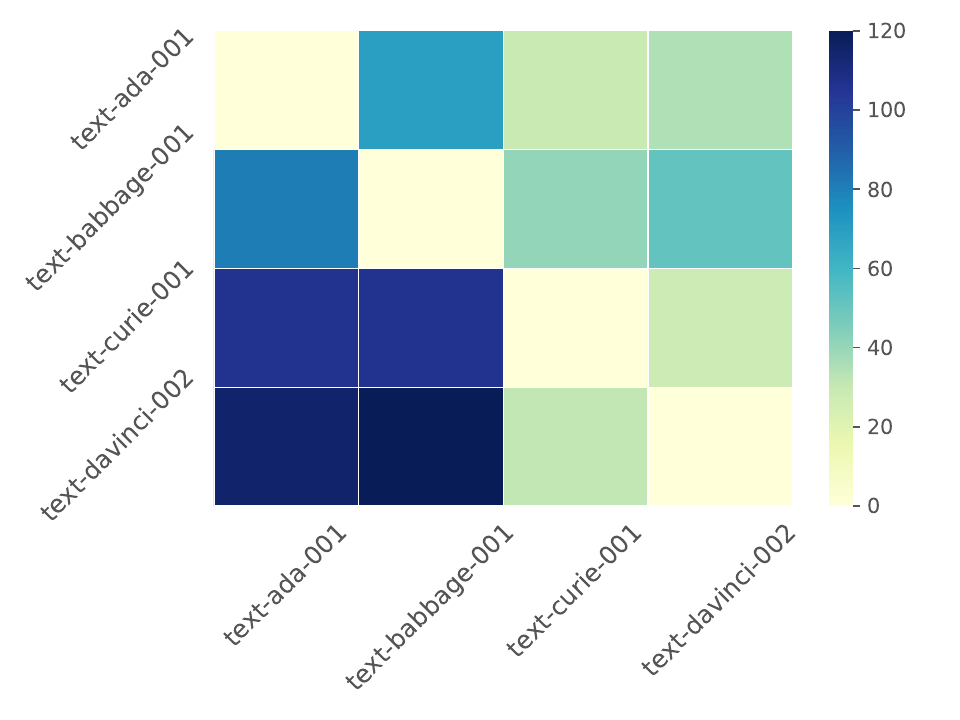}
        \caption{The number of samples that can be answered by one model but not by the other models. Cheaper models can answer many queries that more expensive models cannot.}
        \label{fig:openai_llms}
    \end{minipage}
    \hfill
    \begin{minipage}[b]{0.43\textwidth}
        \includegraphics[width=1.0\linewidth]{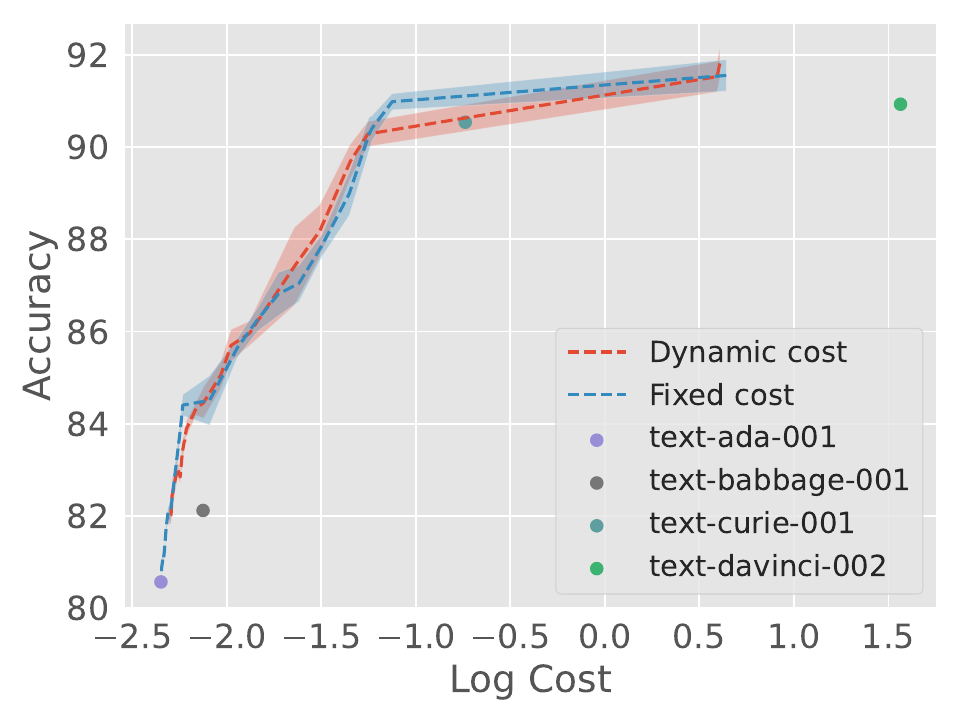}
        \caption{The cost-accuracy trade-off of \ours~with the cost in log scale for better visualization. \ours~with dynamic cost in the reward function slightly decreases the accuracy with a high budget; however, both approaches can perform better than a single LLM.}
        \label{fig:dynamic_cost}
    \end{minipage}
\end{figure}

\subsection{The Heterogeneous Capabilities of Different LLMs}\label{sec:diverse_property}

The fact that \ours~can achieve better accuracy than even the most expensive OpenAI model means that an LLM's performance can vary across queries, and cannot always be determined by its usage cost (or the size of the LLM). In this section, we provide a more rigorous analysis of this observation. 
In Figure~\ref{fig:openai_llms}, the value at position $(i,j)$ is the number of samples that are correctly classified by the $i$-th model but not the $j$-th model. As we can observe, there exist many queries that can be answered by the smaller models, such as \ada~and \babbage, while \curie~and \davinci~give incorrect answers. If \ours~can exploit this heterogeneous relationship between performance and cost to find the most suitable LLM for a query, it can even significantly boost the accuracy of using a single model. Our analysis in Section~\ref{sec:analysis} confirms this hypothesis.

\subsection{Reward Function With Dynamic Cost}

The cost of an LLM is not fixed for every input; for the zero-shot classification task, it primarily depends on the length of the query. We perform an experiment where we use the exact cost of each training input in the reward function~\eqref{formula:reward} and train \ours~with different $p$. 
The results for this dynamic cost setting and the fixed cost are provided in Figure~\ref{fig:dynamic_cost}. As we can observe, using dynamic costs does not lead to significant improvements, although both strategies can achieve better accuracies and lower costs than defaulting to a single LLM. When we set a high budget, dynamic cost setting can even has lower accuracy. 
We hypothesize that using dynamic cost imposes greater penalties on long queries, thereby encouraging MetaLLM to route them to cheaper models; these queries are, however, more complicated and may not be effectively solved by these cheaper models. Therefore, we recommend training \ours~with a fixed cost in the reward function for every training query.

\subsection{Performance of \ours~on Together AI APIs}

\begin{table}[ht!]
\centering

\begin{tabular}{@{}llcc@{}}
\toprule
Setting & Method & Test cost & Test accuracy \\ \midrule
 & \gemma & 866.187 & 64.33 \\
 & \llama & 1559.136 & 76.69 \\
 & \mistral & 1732.374 & 66.68 \\
 & \qwen & 2598.561 & 80.24 \\ \midrule
\multirow{2}{*}{Offline} & \ours~(\llama~budget) & 1474.975 & 75.70\\
 & \ours~(\qwen~budget) & 2287.195 & 79.68 \\ \midrule
\multirow{2}{*}{Online w/o training data} & \ours~(\llama~budget) & 866.187 & 64.33 \\
 & \ours~(\qwen~budget) & 866.187 & 64.33 \\ 
  \midrule
\multirow{2}{*}{Online} & \ours~(\llama~budget) & 1540.663 & 76.87 \\
 & \ours~(\qwen~budget) & 2304.500 & 80.90 \\
\bottomrule
\end{tabular}
\caption{The cost and accuracy of each LLM on Together AI APIs}
\label{tab:together_results}
\end{table}

This section studies the scenario where the LLMs are heterogeneous, i.e., 
the less expensive models can perform better on some queries than the more expensive ones. We perform this experiment using the LLMs provided by Together AI and the MMLU dataset. Table~\ref{tab:together_results} provides the accuracy and the cost of each approach, including defaulting to the same LLM (the first four rows) and our \ours. As we can observe, when defaulting to a single LLM, a more expensive option does not guarantee a better performance; for example, \mistral~is more expensive than \llama~but yields lower performance.

On the other hand, training \ours~with $p=0$ yields the best performance ($80.90\%$ accuracy) on MMLU. 
Since the distribution of the auxiliary training set is far from the test set, optimizing the reward function does not provide a good policy, i.e., obtaining the highest accuracy. On the other hand, without the initialization step, the multi-arm bandit algorithm only defaults to \gemma.
Setting a positive value to $p$ decreases the cost substantially while having a minor degradation in the accuracy. 
Notably, \ours~ can achieve better performance than the second-best LLM, \llama, but with a lower cost. These results further confirm that \ours~provides a cost-efficient yet high-performance policy. 

\subsection{Analysis of \ours's Cost Scaling Parameter}\label{sec:analysis}
\begin{figure}[ht!]
\begin{minipage}[b]{0.48\linewidth}
    \centering
    \includegraphics[width=\textwidth]{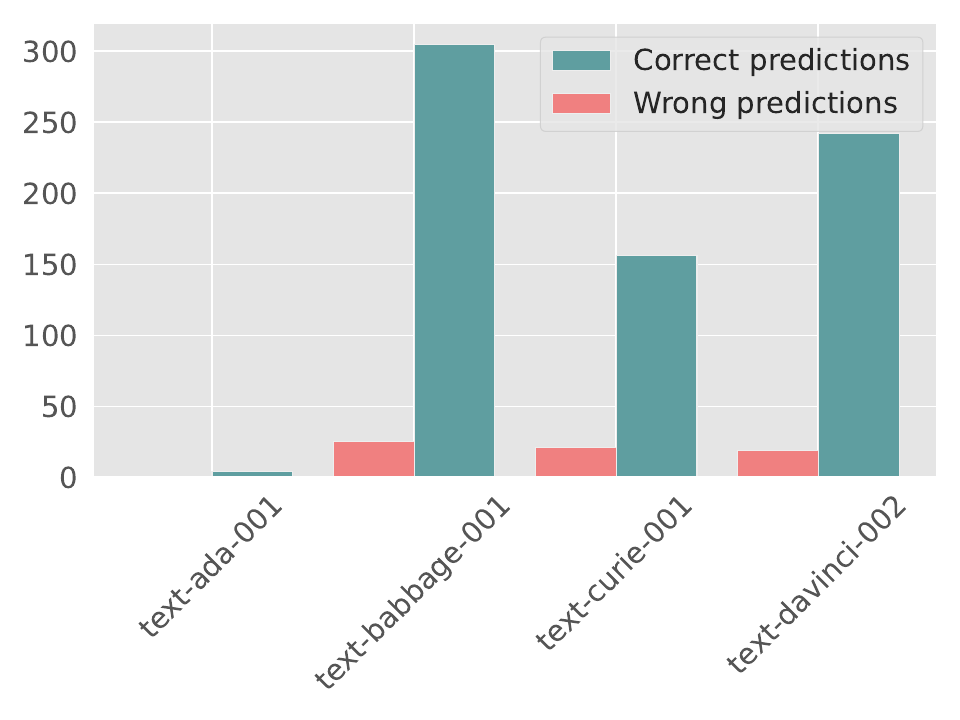}
    \caption{The histogram of OpenAI LLMs selected by \ours~with $p=0$.}
    \label{fig:open_pred_best}
\end{minipage}\hfill
\begin{minipage}[b]{0.48\linewidth}
        \centering
    \includegraphics[width=\textwidth]{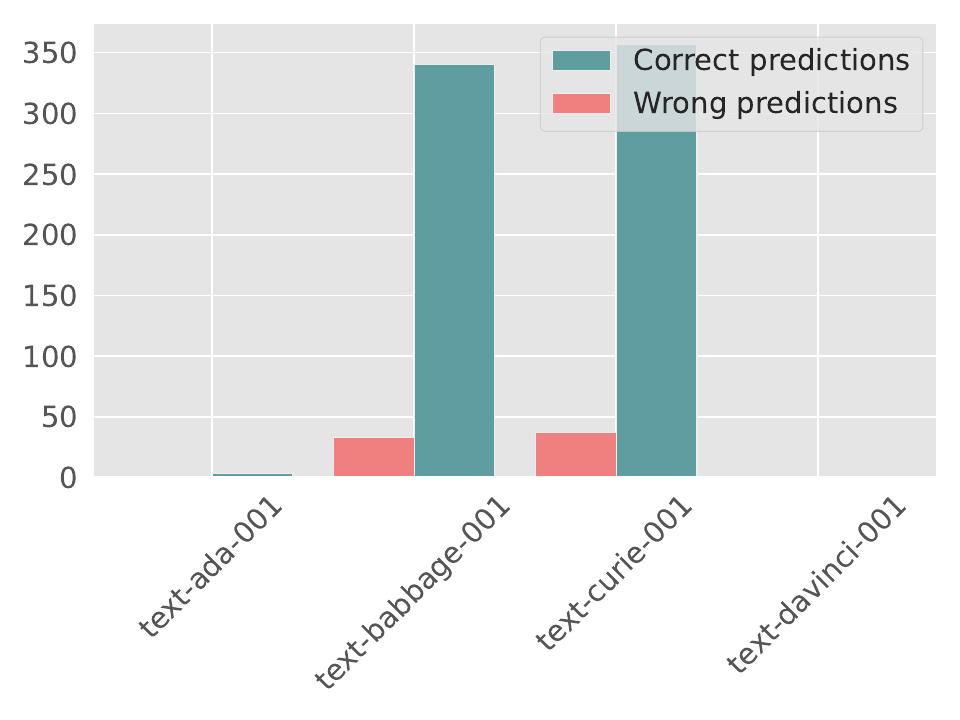}
    \caption{The histogram of OpenAI LLMs selected by \ours~with $p=0.001$.}
    \label{fig:openai_pred_eff}
\end{minipage}
\end{figure}

In this section, we study the characteristics of \ours~with different values of $p$ in the reward function. First, we consider the scenario where the user only optimizes for the performance (i.e., $p=0$). Figure~\ref{fig:open_pred_best} shows the histogram or the frequency each OpenAI's LLM is selected by \ours, separated by the number of correct and incorrect predictions, in the SST-2 dataset. As we can observe, \ours~indeed can learn whether an LLM can accurately answer a query. As discussed in Section~\ref{sec:diverse_property}, there exist many queries that can be correctly answered by less expensive models such as \babbage, and \ours~chooses \babbage~more often than choosing \curie~or \davinci~. 

Figure~\ref{fig:openai_pred_eff} shows the histogram when putting a small regularization $p=0.001$ on the cost component in the reward function. In this scenario, \ours~rarely picks \davinci, while still being able to achieve $90.98\%$ accuracy; as a reminder, this is still better than defaulting to the most expensive model, as previously observed in Table~\ref{tab:openai_results}. 



\section{Conclusion}
In this paper, we study the problem of dynamically selecting an LLM out of a set of LLMs for an input that achieves optimal performance and cost efficiency. To solve this problem, we propose a multi-armed bandit framework that can learn the reward of querying the correct model at a low price. 
Our approach, denoted as \ours, is lightweight and applicable to any set of off-the-shelf LLMs and thus is versatile in practical use cases. Empirical results show that \ours~can improve the accuracy of the best API by around $1\%$ while significantly reducing the cost of zero-shot text classification and multiple-choice question answering tasks by up to $60\%$.


\section{Limitations and Societal Impacts}\label{sec:limitation_impact}
\subsection{Limitations} As mentioned in the previous sections, we only study \ours~'s framework on zero-shot text classification tasks as these are important in NLP applications and increasingly utilize LLMs as the base predictors. Another reason is that it is straightforward to determine the correct LLM outputs and set up the reward function accordingly. As our paper aims to to demonstrate the potential of the \ours's framework, this is sufficient.

However, the \ours{} framework can be extended to arbitrary language tasks, such as as question answering or text generation, by modifying the reward function to incorporate suitable metrics assessing the quality of the responses. Due to the complexities of designing such reward function, these directions deserve independent studies. We leave them to future work.



\ours{} also only trains a simple linear model whose input is the extracted feature of the query, which can ignore more fine-grained features. Building a more complex reward model and utilizing other information from the query, such as the domain of the input and the demand of the user, may further facilitate better the needs of the applications and improve the performance of \ours.

Finally, we optimize \ours~with two values in the reward function: the performance and the cost of querying the API. However, several aspects to evaluate the model in practice could be incorporated into the reward, such as the inference time, the robustness of the model, emergent abilities, or even the information on the training distribution. Combining those factors can help build a more powerful and reliable AI system for diverse purposes.

\subsection{Societal Impacts}
Large language models, with their emergent abilities, have transformed our lives by assisting in several tasks. However, large models come with high inference costs; therefore they are very expensive to deploy and may cause harmful effects to the environment by high power consumption. Our framework helps reduce the cost of querying LLMs substantially by routing the input to cheaper models that may return the correct answer and can even improve performance by utilizing the combination of many LLMs. \ours~is applicable to any set of off-the-shelf LLMs, being useful for future AI systems with more modern language models.

\bibliography{ref}
\bibliographystyle{colm2025_conference}


\end{document}